\documentclass[Afour,sageh,times]{sagej}

\usepackage{moreverb,url}

\usepackage[colorlinks,bookmarksopen,bookmarksnumbered,citecolor=red,urlcolor=red]{hyperref}

\newcommand\BibTeX{{\rmfamily B\kern-.05em \textsc{i\kern-.025em b}\kern-.08em
T\kern-.1667em\lower.7ex\hbox{E}\kern-.125emX}}

\def\volumeyear{2020}

\begin{document}


\title{Detection of Genuine and Posed Facial Expressions of Emotion: A Review}

\author{Shan Jia\affilnum{1}, Shuo Wang\affilnum{2}, Chuanbo Hu\affilnum{3}, Paula Webster\affilnum{2}, Xin Li\affilnum{3}}

\affiliation{\affilnum{1}State Key Laboratory of Information Engineering in Surveying Mapping and Remote Sensing, Wuhan University, Wuhan, China\\
\affilnum{2}Department of Chemical and Biomedical Engineering, West Virginia University, Morgantown, WV, USA\\
\affilnum{3}Lane Department of Computer Science and Electrical Engineering, West Virginia University, Morgantown, WV, USA}
\corrauth{Xin Li, Lane Department of Computer Science and Electrical Engineering, West Virginia University, Morgantown, WV, USA}

\email{xin.li@mail.wvu.edu}

\begin{abstract}
Facial expressions of emotion play an important role in human social interactions. However, posed acting is not always the same as genuine feeling. Therefore, the credibility assessment of facial expressions, namely, the discrimination of genuine (spontaneous) expressions from posed(deliberate/volitional/deceptive) ones, is a crucial yet challenging task in facial expression understanding. Rapid progress has been made in recent years for automatic detection of genuine and posed facial expressions. This paper presents a general review of the relevant research, including several spontaneous vs. posed (SVP) facial expression databases and various computer vision based detection methods. In addition, a variety of factors that will influence the performance of SVP detection methods are discussed along with open issues and technical challenges.

\end{abstract}


\maketitle

\section{Introduction}
Facial expressions, one of the main channels for understanding and interpreting emotions among social interactions, have been studied extensively in the past decades \citep{motley1988facial,zuckerman1976encoding}. Most existing research works have focused on automatic facial expression recognition based on Ekman’s theories~\citep{ekman1997universal}, which suggests six basic emotions universal in all cultures, including happiness, surprise, anger, sadness, fear, and disgust. 
However, are facial expressions always the mirror of our innermost emotions as we have believed for centuries? Recent research~\citep{crivelli2015smiles} has found that facial expressions do not always reflect our true feelings. Instead of reliable readouts of people's emotional states, facial expressions tend to be increasingly posed and even deliberately to show our intentions and social goals. Therefore, understanding the credibility of facial expressions in revealing emotions has become an important yet challenging task in human behavioral research especially among the studies of social interaction, communication, anthropology, personality, and child development~\citep{bartlett1999measuring}. 

In the early years, research of discriminating genuine facial expressions from posed ones heavily relied on a variety of observer-based systems~\citep{mehu2012reliable}. Rapid advances in computer vision and pattern recognition especially deep learning techniques have recently opened up new opportunities for automatic and efficient separation of genuine facial expressions from posed ones. A variety of SVP facial expression detection methods ~\citep{valstar2006spontaneous, dibeklioglu2010eyes, wu2014spontaneous, huynh2017discrimination, park2020differences}, as well as publicly available databases~\citep{wang2010natural, mavadati2016extended, cheng20184dfab,pfister2011differentiating}, have been proposed for facial expression credibility analysis.

As of today, there has been no systematic survey yet to summarize the advances of SVP facial expression detection in the past two decades. To fill in this gap, we present a general review of the pioneering works as well as most recent studies in this field including both existing SVP databases and automatic detection algorithms. Through literature survey and analysis, we have organized existing SVP detection methods into four categories (action units, spatial patterns, visual features, and hybrid) and identified a number of factors that will influence the performance of SVP detection methods. Furthermore, we attempt to provide some new insights into the remaining challenges and open issues to address in the future.

\section{Spontaneous vs. posed facial expression databases}
Early studies on facial expressions are mostly based on posed expressions due to the easier collection process, where the subjects are asked to display or imitate each basic emotional expression. Spontaneous facial expressions, however, as natural expressions, need to be induced by various stimuli, such as odours~\citep{simons2003disturbance}, photos~\citep{gajvsek2009multi}, and video clips~\citep{pfister2011differentiating, petridis2013mahnob}. There have been several databases with single or multiple facial expressions collected to promote the research in automatic facial expression credibility detection. This section focuses on databases with both spontaneous and posed facial expressions, and provides the details of existing public databases (see an overview in Table \ref{tab:01}).

The MMI facial expression database~\citep{pantic2005web} was first collected with only posed expressions for facial expression recognition. Later data with three spontaneous expressions (disgust, happiness, and surprise) were added with audio-visual recordings based on video clips as stimulus~\citep{valstar2010induced}. USTC-NVIE~\citep{wang2010natural} is a visible and infrared thermal SVP database. Six spontaneous emotions consisting of image sequences from onset to apex\footnote[1]{Onset, apex, along with offset, and neutral, are four possible temporal segments of facial actions during the expression development (generally in the order of neutral$\rightarrow$ onset$\rightarrow$ apex$\rightarrow$ offset$\rightarrow$ neutral). In the onset phase, muscles are contracting and changes in appearance are growing stronger. In the apex phase, the facial action is at a peak with no more changes in appearance. The offset phase describes that the muscles of the facial action are relaxing and the face returns to its original and neutral appearance, where there are no signs of activation of the investigated facial action.}, were also induced by screening carefully selected videos, while the posed emotions consist of apex images.
CK+ database~\citep{lucey2010extended}, UvA-NEMO~\citep{dibekliouglu2012you}, and MAHNOB database~\citep{petridis2013mahnob} all focused on the smile, which is the easiest emotional facial expression to pose voluntarily. Specifically, the video sequences in CK+ database were fully coded based on Facial Action Coding System (FACS) \citep{ekman1997face} for facial action units (AUs) as emotion labels, while videos in MAHNOB recorded both smiles and laughter with microphones, visible and thermal cameras.

\begin{table*}[t]
\newcommand{\tabincell}[2]{\begin{tabular}{@{}#1@{}}#2\end{tabular}} %
  \centering
\footnotesize
  \caption{Description of SVP facial expression databases}
    \begin{tabular}{p{1.55cm}p{1.02cm}llllllp{2.2cm}}
    \hline
    \textbf{Dataset}  & \textbf{Expression} &{\textbf{\#Sub}} & \textbf{\#M/F} & \textbf{Age} & \textbf{\#P/S} & \textbf{Format} & \textbf{Feature} & {\textbf{Reference}} \\
    \hline
    MMI    & Multiple & 25    & 13/12 & 20-32 & 2489/392 & Video & \tabincell{l}{Audio-visual; single and\\ combinations of AUs} & {\cite{valstar2010induced}} \\
    USTC-NVIE  & Multiple & 215   & 157/58 & 17-31 & -/-   & Frame & Visible + infrared thermal images & {\cite{wang2010natural}} \\
    CK+     & {Smile} & 210   & {65/145} & {18-50} & {593/122} & Frame& \tabincell{l}{Multiple posed expressions, \\only un-posed smile, FACS coded} & {\cite{lucey2010extended}} \\
    SPOS Corpus  & Multiple & 7     & 4//3  & /     & 51/147 & Frame  & Visible + infrared & {\cite{pfister2011differentiating}} \\
    UvA-NEMO  & Smile & 400   & 215/185 & 8-76 & 643/597 & Video & The largest smile database &{\cite{dibekliouglu2012you}} \\
     MAHNOB   & Smile & 22    & 12/10 &$\sim$28   & 563/101 & Video & Audio-visual, thermal recording &{\cite{petridis2013mahnob}} \\
    BioVid   & Pain  & 90    &{45/45} & 18-65 & 630/8700 & Video & \tabincell{l}{Biopotential signals,\\ depth information} &\cite{walter2013biovid} \\
    DISFA  & Multiple & 27    & 15/12 & 18-50 & 0/54  & Video & \tabincell{l}{AU labels and landmarks} &{\cite{mavadati2013disfa}} \\
    DISFA+   & Multiple & 9     & 4/5  & 18-50 & 644/0 & Frame  &  AU labels, 42 facial actions &{\cite{mavadati2016extended}} \\
    SASE-FE	&Multiple&	50&	-/-	&19-36	&300/300&	Video&	3 subsets&	\cite{wan2017results}\\
    4DFAB   & Multiple  & 180   & 120/60 & 5-75 & -/-   & 4D video & \tabincell{l}{Dynamic high-resolution 3D faces, \\79 face landmarks} & \cite{cheng20184dfab} \\
    \hline
    \end{tabular}%
  \label{tab:01}%
\end{table*}%

\begin{figure*}[h!]
\begin{center}
\includegraphics[width=16.6cm]{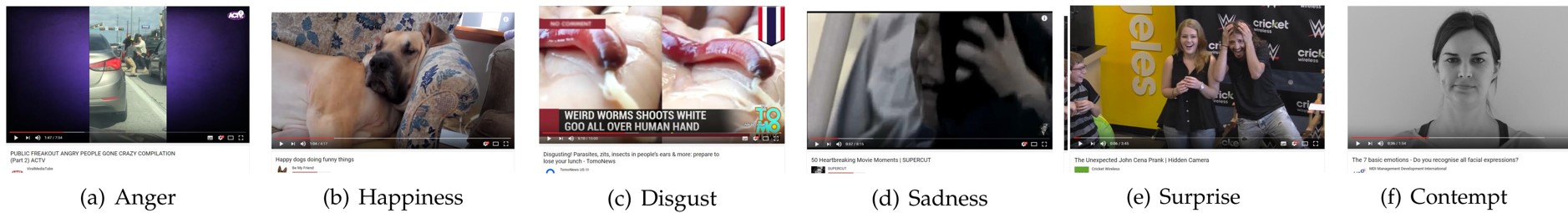}
\end{center}
\caption{Examples of video clips to induce specific emotions in SASE-FE database (Copyright permission is obtained from~\cite{kulkarni2018automatic}).}
\label{fig:02}
\end{figure*}

SPOS Corpus~\citep{pfister2011differentiating} included six basic SVP emotions, with labels for onset, apex, offset and end by two annotators according to subjects' self-reported emotions. BioVid dataset~\citep{walter2013biovid} specifically targeted pain with heat stimulation, and both biosignals (such as SCL, ECG, EMG, and EEG) and video signals were recorded. DISFA and DISFA+ database~\citep{mavadati2013disfa, mavadati2016extended} contain sponaneous and posed facial expressions respectively, with 12 coded AUs labels by FACS and 66 landmark points. In addition to basic facial expressions, DISFA+ also includes 30 facial actions by asking participants to imitate  and pose the specific action. Proposed for ChaLearn LAP Real Versus Fake Expressed Emotion Challenge in 2017, the SASE-FE database~\citep{wan2017results, kulkarni2018automatic} collected 6 expressions by asking participants to pose artificial facial expressions or showing participants video clips to induce genuine ones. Figure \ref{fig:02} illustrates several examples of video clips selected by psychologists to induce specific emotions in this database.

Most recently, a large scale 4D database,  4DFAB~\citep{cheng20184dfab}, was introduced with 6 basic SVP expressions, recorded in four different sessions spanning over a five-year period. This is the first work to investigate on the use of 4D spontaneous behaviours in biometric applications.

\vspace{-0.2cm}
\section{Detection of genuine and posed facial expressions}
Posed facial expressions, due to the deliberate and artificial nature, always differ from genuine ones remarkably in terms of intensity, configuration, and duration, which have been explored as distinct features for SVP facial expression recognition. Based on different distinct clues, we classify existing methods into four categories: {\it muscle movement (action units) based, spatial patterns based, visual features based, and hybrid methods}. 
\vspace{0.1cm}
\subsection{Muscle movement (action units) based}
\begin{figure*}[h!]
\begin{center}
\includegraphics[width=13.5cm]{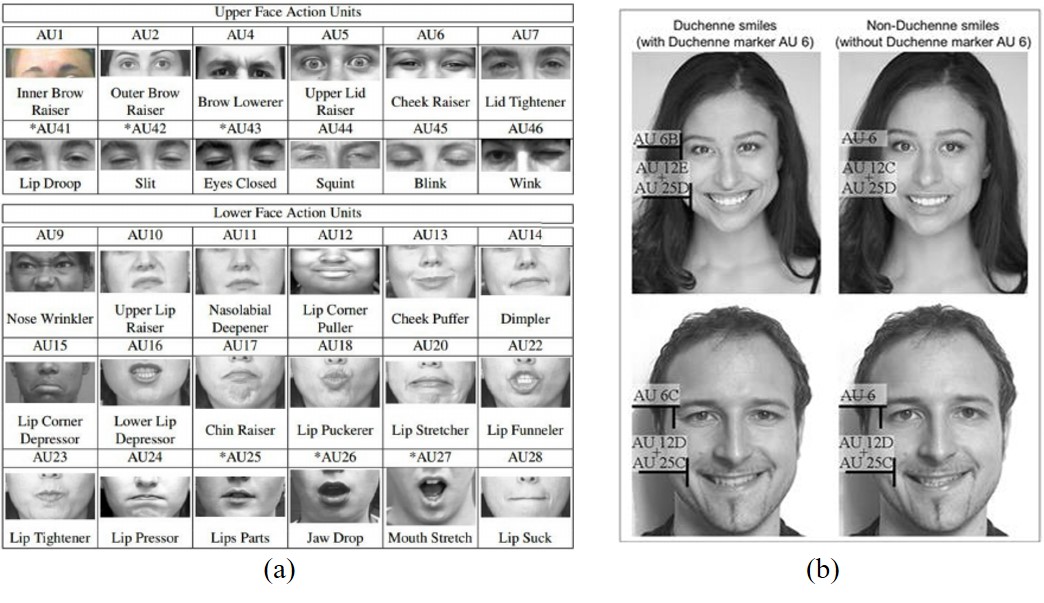}
\end{center}
\caption{Examples of FACS AUs, (a) Upper and lower face AUs (Copyright permission is obtained from~\cite{la2015intraface}), (b) Different AUs in Duchenne smiles (AU 6, 12, 25) and non-Duchenne smiles (AU12, 25) (Copyright permission is obtained from~\cite{bogodistov2017proximity}).}
\label{fig:03}
\end{figure*}

Early research on distinguishing genuine facial expressions from posed ones rely a lot on the analysis of facial muscle movement. This class of methods are based on the assumption that some specific facial muscles are particularly trustworthy cues due to the intrinsic difficulty of producing them voluntarily~\citep{ekman2003darwin}. In these studies, the Facial Action Coding System (FACS)~\citep{ekman2005face} is the most widely-used tool for decomposing facial expressions into individual components of muscle movements, called Action Units (AUs), as shown in Figure \ref{fig:03}(a). Several studies have explored the differences of muscle movements (AUs) in spontaneous and posed facial expressions, including the AUs amplitude, maximum speed, and duration (please refer to Figure \ref{fig:03}(b) for an example).

It is known that spontaneous smiles have a smaller amplitude, but a larger and more consistent relation between amplitude and duration than deliberate smiles. Based on this observation, method in \citep{cohn2003timing} used timing and amplitude measures of smile onsets for detection and achieved a 93\% recognition rate with a linear discriminant analysis classifier (LDA). The method in~\citep{valstar2006spontaneous} was the first attempt to automatically determine whether an observed facial action was displayed deliberately or spontaneously. They proposed to detect SVP brow actions based on automatic detection of three AUs (AU1, AU2, and AU4) and their temporal segments (onset, apex, offset) produced by movements of the eyebrows. Experiments on combined databases achieved 98.80\% accuracy. Later works ~\citep{bartlett2006automatic,bartlett2008data} extracted five statistic features (median, maximum, range, first to third quartile difference) of 20 AUs in each video segment for classification of posed and spontaneous pain. They reported a 72\% classification accuracy on their own dataset. To detect SVP smile, method in \citep{schmidt2009comparison} quantified lip corner and eyebrow movement during periods of visible smiles and eyebrow raises, and found maximum speed and amplitude were greater and duration shorter in deliberate compared to spontaneous eyebrow raises. Aiming at multiple facial expressions, the method~\citep{saxen2017real} generated a 440-dimensional statistic feature space from the intensity series of 7 facial AUs, and increased the performance to 73\% by training an ensemble of Rank SVMs on SASE-FE database. Differently, recent work in \citep{racovicteanu2019spontaneous} used AlexNet CNN architecture on 12 AU intensities to obtain the features in transfer learning task. Training on DISFA database, and testing on SPOS, the method achieved an average accuracy of 72.10\%. A brief overview of these methods has been shown in Table \ref{tab:02}.

\begin{table*}[ht]
\newcommand{\tabincell}[2]{\begin{tabular}{@{}#1@{}}#2\end{tabular}} 
  \centering
\footnotesize
  \caption{A brief overview of muscle movement based SVP detection methods.}
    \begin{tabular}{p{1.8cm}p{4.2cm}lp{2.5cm}p{1.8cm}p{1.57cm}l}
    \hline
    \textbf{Reference} & \textbf{Method (features)} & \textbf{Expression} & \textbf{AU} & \textbf{Classification} & \textbf{Database} & \multicolumn{1}{p{4.045em}}{\textbf{Accuracy}} \\
\hline
    \cite{cohn2003timing} & Using timing and amplitude measures of smile onsets & Smile & 6, 12, 15, 17 & LDA & Self-collected & {93.00\%} \\
    \cite{valstar2006spontaneous} & Temporal dynamics of brow actions based on AUs and their temporal segments (onset, apex, offset)  & Multiple (6) & 1, 2, 4 &Relevance Vector Machine & MMI+DS118+ CK+(262)  & {90.80\%} \\
    \cite{bartlett2008data} & Statistic features of 20 AUs in each video segment   & Pain  & 1, 2, 4-7, 9, 10, 12, 14, 15, 17, 18, 20, 23-26 & Nonlinear SVM & Self-collected & {72.00\%} \\
    \cite{schmidt2009comparison} & Maximum speed and amplitude of movement onset of lip corner and eyebrow; AFIA to measure movement & Smile  & 6, 12, 14, 15, 17, 23, 24, 50 & /     & Self-collected & / \\
    \cite{saxen2017real} & statistic features (440-dimensional) from the intensity time series of 7 facial AUs & Multiple (6) & 1, 2, 4, 6, 9, 12, 25 & Rank SVMs & SASE-FE & 73.00\% \\
    \cite{racovicteanu2019spontaneous} & AlexNet CNN architecture on 12 AU intensities to obtain the features in a tranfer learning manner & Multiple (6) & 1, 2, 4-6, 9, 12, 15, 17, 20, 25, 26 & SVM   & DISFA, SPOS & 72.10\% \\
\hline
    \end{tabular}%
  \label{tab:02}%
\end{table*}%

\subsection{Spatial patterns based}
This category of methods aim at exploring spatial patterns based on temporal dynamics of different modalities such as facial landmarks and shapes of facial components. A multimodal system based on fusion of temporal attributes including tracked points of face, head and shoulder was proposed in \citep{valstar2007distinguish} to discern posed from spontaneous smiles. Best results were obtained with late fusion of all modalities of 94\% on 202 videos from MMI database. Specifically regarding smile, a study in \citep{van2008tooth} analyzed differences in tooth display, lip-line height, and smile width between SVP smiles. They revealed several findings in SVP smiling differences. For example, maxillary lip-line heights in genuine smiles were significantly higher than those in posed smiles. When compared to genuine smiling, the tooth display in the (pre)molar area of posed smiling decreased by up to 30\%, along with a significant reduction of smile width. Spatial patterns based on distance and angular features for eyelid movements were used in \citep{dibeklioglu2010eyes} and achieved 85\% and 91\% accuracy in discriminating between SVP smiles on the BBC and CK databases, respectively. Based on fusing dynamics signals of eyelids, cheeks, and lip corners, more recent methods \citep{dibekliouglu2012you, dibekliouglu2015recognition} achieved promising detection results on several SVP smile databases. 

In multiple SVP facial expression detection, different schemes for spatial pattern modeling, including Restricted Boltzmann Machines (RBMs) based, were studied in ~\citep{wang2015posed, wang2016capturing}, Latent Regression Bayesian Network based in~\citep{gan2017differentiating}, and interval temporal restricted Boltzmann machine (IT-RBM) in~\citep{wang2019novel}. Results on several SVP databases confirmed the discriminative power and reliability of spatial patterns in distinguishing genuine and posed facial expressions. Similarly, \citep{huynh2017discrimination} used mirror neuron modeling and Long-short Term Memory (LSTM) \cite{hochreiter1997long} with parametric bias to extract features in the spatial-temporal domain from extracted facial landmarks, and achieved 66\% accuracy on the BABE-FE database. Table \ref{tab:03} presents an overview of these spatial pattern based detection methods. 

\begin{table*}[htbp]
\newcommand{\tabincell}[2]{\begin{tabular}{@{}#1@{}}#2\end{tabular}} 
  \centering
\footnotesize
  \caption{A brief overview of spatial patterns based SVP detection methods.}
    \begin{tabular}{p{2.85cm}p{6.2cm}lp{1.4cm}p{1.62cm}p{1.15cm}}
\hline
    \textbf{Reference} & \textbf{Method (features)} & \textbf{Expression} & \textbf{Classification} & \textbf{Database} & \textbf{Accuracy} \\
\hline
    \cite{valstar2007distinguish} & \tabincell{l}{Fusing temporal dynamics of head (6 features), face (12\\ points), and shoulder (5 points) modalities} & Smile & \tabincell{l}{GentleSVM-\\Sigmoid} & MMI (202) & {94.00\%} \\
    \cite{van2008tooth} & \tabincell{l}{Analyzing tooth display, lip position and smile width in\\ a dental perspective} & Smile & /     & Self-collected & / \\
    \cite{dibeklioglu2010eyes} & \tabincell{l}{Distance-based and angular features for eyelid\\ movements} & Smile & Naive Bayes & \tabincell{l}{BBC,\\ CK} & \tabincell{l}{85.00\%; \\91.00\%} \\
    \cite{dibekliouglu2012you}& \tabincell{l}{Fusing the dynamics of eyelid, cheek, and lip corner\\ movements} & Smile & linear SVM & \tabincell{l}{BBC,\\ SPOS,\\ UvA-NEMO}  & \tabincell{l}{90.00\%,\\ 75.00\%,\\ 87.02\%} \\
    \cite{dibekliouglu2015recognition} & Dynamics of eyelid, cheek, and lip corner movements & Smile & SVM   & \tabincell{l}{BBC,\\ SPOS,\\ UvA-NEMO, \\MMI} & \tabincell{l}{90.00\%, \\78.75\%, \\92.10\%, \\89.69\%} \\
    \cite{wang2015posed, wang2016capturing} & \tabincell{l}{Spatial pattern modeling based on multiple RBMs and\\ incorporating gender and expression categories as\\ privileged information}  & Multiple (6) & RBMs  & \tabincell{l}{SPOS,\\ USTC-NVIE,\\ MMI} & \tabincell{l}{76.07\%,\\ 92.61\%,\\ 89.79\%} \\
    \cite{gan2017differentiating} & \tabincell{l}{Spatial patterns based on Latent Regression Bayesian\\ Network from he displacements of facial feature points} & Multiple (6) &\tabincell{l}{ Bayesian\\ Networks} & \tabincell{l}{SPOS,\\ USTC-NVIE}  & \tabincell{l}{76.07\%, \\98.74\%} \\
    \cite{huynh2017discrimination} & \tabincell{l}{Spatial-temporal features using mirror neuron modeling\\ and LSTM with parametric bias from facial landmarks} & Multiple (6) & \tabincell{l}{Gradient\\ boosting}  & SASE-FE & \multicolumn{1}{l}{66.70\%} \\
    \cite{wang2019novel} & \tabincell{l}{Universal spatial patterns and complicated temporal\\ patterns using IT-RBM dynamic model} & Multiple (6) & \tabincell{l}{Bayesian\\network} & \tabincell{l}{SPOS,\\ DISFA+} & \tabincell{l}{83.76\%,\\ 96.24\%}  \\
\hline
    \end{tabular}%
  \label{tab:03}%
\end{table*}%

\subsection{Visual features based}
Visual features (appearance) based such as \citep{littlewort2009automatic} designed a two-stage system to distinguish faked pain from real pain. It consisted of a detection stage for 20 facial actions using Gabor features and a SVM classification stage. The two-stage system achieved 88\% accuracy on the UvA-NEMO dataset. Another method \citep{pfister2011differentiating} proposed a new feature, named Completed local binary patterns from Three Orthogonal Planes (CLBP-TOP), and fused the NIR and VIS modalities with Multiple Kernel Learning (MKL) classifier, which achieved outstanding detection performance of 80.0\% on the SPOS database. Also based on infrared images, \citep{liu2012posed} used facial temperature information from thermal images, and extracted statistical features from five facial subregions for SVP facial expression detection. Finally, the approach in \citep{gan2015posed} proposed to use pixel-wise difference between onset and apex face images as input features of a two-layer deep Boltzmann machine to distinguish SVP expressions. They achieved 84.62\% and 91.73\% on the SPOS and USTC-NVIE databases respectively.

More recently, \cite{mandal2016distinguishing} explored several features, including deep CNN features, local phase quantization (LPQ), dense optical flow and histogram of gradient (HOG), to classify SVP smiles. With Eulerian Video Magnification (EVM) for micro-expression smile amplification, the HOG features outperformed other features with an accuracy of 78.14\% on UvA-NEMO Smile Database. Instead of using pixel-level differences, the method~\citep{xu2017convolutional} designed a new layer named “comparison layer” for deep CNN to generate high-level representations of the differences of onset and apex images, and verified its effectiveness on SPOS (83.34\%) and USTC-NVIE database (97.98\%). The latest work \cite{tavakolian2019learning} present a Residual Generative Adversarial Network (R-GAN) based method to discriminate SVP pain expression by magnifying the subtle changes in faces. Experiment results have shown the state-of-the-art performance on three databases, with 91.34\% on UNBC-McMaster~\citep{lucey2011painful} with  spontaneous pain expressions only, 85.05\% on BiodVid, and 96.52\% on STOIC~\citep{roy2007stoic} with posed expressions only. A brief overview of these methods has been shown in Table \ref{tab:04}.
\begin{table*}[h]
\newcommand{\tabincell}[2]{\begin{tabular}{@{}#1@{}}#2\end{tabular}} 
  \centering
\footnotesize
  \caption{A brief overview of visual features based SVP detection methods.}
    \begin{tabular}{p{2.65cm}p{5.84cm}lp{1.75cm}p{2cm}p{1.05cm}}
\hline
    \textbf{Reference} & \textbf{Method (features)} & \textbf{Expression} & \textbf{Classification} & \textbf{Database} & \textbf{Accuracy} \\
\hline
    \cite{littlewort2009automatic} &  Gabor features based & Pain  & Gaussian SVM & UvA-NEMO &{88.00\%} \\
    \cite{pfister2011differentiating} & Spatiotemporal local texture descriptor (CLBP-TOP), fusing the NIR and VIS modalities & Multiple (6) & MKL   & SPOS  & {80.00\%} \\
    \cite{liu2012posed} & \tabincell{l}{Temperature features from Infrared thermal images} & Multiple (6) & \tabincell{l}{Bayesian\\ Networks} & USTC-NIVE & {76.70\%} \\
    \cite{gan2015posed} & \tabincell{l}{A two-layer deep Boltzmann machine model based}  & Multiple (6) & Haarcascades & \tabincell{l}{SPOS,\\USTC-NVIE} &\tabincell{l}{ 84.62\%, \\91.73\%} \\
    \cite{mandal2016distinguishing} & \tabincell{l}{Several features: using CNN face features, LPQ,\\ dense optical flow and HOG, and HOG with the\\ best result} & Smile & Linear SVM & UvA-NEMO & {78.14\%} \\
    \cite{xu2017convolutional} & \tabincell{l}{Learned features based on CNN from difference\\ image of onset and apex images} & Multiple (6) & Linear SVM & \tabincell{l}{SPOS,\\ USTC-NVIE} & \tabincell{l}{83.34\%,\\ 97.98\%}\\
    \cite{tavakolian2019learning} & \tabincell{l}{Encoding the dynamic and appearance of a video\\ into an image map based on spatiotemporal pooling,\\ then using R-GAN model for discrimination} & Pain  & Softmax & \tabincell{l}{BioVid Heat Pain, \\STOIC, \\UNBC-McMaster} &\tabincell{l}{ 85.05\%, \\96.52\%, \\ 91.34\% }\\
\hline
    \end{tabular}%
  \label{tab:04}%
\end{table*}%

\begin{table*}[tbp]
\newcommand{\tabincell}[2]{\begin{tabular}{@{}#1@{}}#2\end{tabular}} 
  \centering
\footnotesize
  \caption{A brief overview of hybrid methods for SVP detection.}
    \begin{tabular}{p{2.48cm}p{6.73cm}llp{1.6cm}p{1.0cm}}
\hline
    \textbf{Reference} & \textbf{Method (features)}& \textbf{Expression} & \textbf{Classification} & \textbf{Database} & {\textbf{Accuracy}} \\
\hline
    \cite{zhang2011geometry} & SIFT appearance based features and FAP geometric features & Multiple (6) & RBF SVM & USTC-NVIE & 79.40\% \\
    \cite{li2017combining} & \tabincell{l}{Combining sequential geometric features based on facial\\ landmarks and texture features using HOG} & Multiple (6) & Sigmoid & SASE-FE & 68\% \\
    \cite{mandal2017spontaneous} & \tabincell{l}{Fusing subtle (micro) changes by tracking a series of facial\\ fiducial markers with local and gobal motion based on dense\\ optical flow} & Smile & SVM   & UvA-NEMO & 74.68\% \\
    \cite{kulkarni2018automatic} & \tabincell{l}{Combining learned static CNN representations from still\\ images with facial landmark trajectories} & Multiple (6)& Linear SVM  & SASE-FE & 70.20\% \\
    \cite{saito2020classification} & \tabincell{l}{Combining hardware (16 sensors embedded with the smart\\ eyewear) with software-based method to get geometric\\ and temporal features} & Smile & Linear SVM & Self-collected & 94.60\%\\
\hline
    \end{tabular}%
  \label{tab:05}%
\end{table*}%

\vspace{0.2cm}
\subsection{Hybrid methods}
Hybrid methods combined different classes of features for discriminating SVP facial expressions. Experiments on still images were conducted in \citep{zhang2011geometry} to show that appearance features (e.g., Scale-Invariant Feature Transform (SIFT) \cite{lowe2004distinctive}) play a significantly more important role than geometric features (e.g., facial animation parameters (FAP) \cite{aleksic2006automatic}) on SVP emotion discrimination, and fusion of them leads to marginal improvement over SIFT appearance features. The average classification accuracy of six emotions is 79.4\% (surprise achieved the best result of 83.4\% while anger with the worst of 77.2\%) on the USTC-NVIE database. Sequential geometric features based on facial landmarks and texture features using HOG were combined in~\citep{li2017combining}. A temporal attention gated model is designed for HOG features, combining with LSTM autoencoder (eLSTM) to capture discriminative features from facial landmark sequences. The proposed model performed well on most emotions on SASE-FE database, with an average accuracy of 68\%. \citep{mandal2017spontaneous} fused subtle (micro) changes by tracking a series of facial fiducial markers with local and global motion based on dense optical flow, and achieved 74.68\% using combined features from eyes and lips, slightly better than using only the lips (with 73.44\%) and using only the eyes (with 71.14\%) on the UvA-NEMO smile database. A different hybrid method in \citep{kulkarni2018automatic} combined learned static CNN representations from still images with facial landmark trajectories, and achieved promising performance not only in emotion recognition, but also in detecting genuine and posed facial expressions on the BABE-FE database with data augmentation (70.2\% accuracy). 
Most recently, \citep{saito2020classification} combined hardware (16 sensors embedded with the smart eye-wear) with software-based method to get geometric and temporal features to classify smiles into either ‘spontaneous’ or ‘posed’, with an accuracy of 94.6\% on their own database. See Table \ref{tab:05} for a brief summary of these hybrid SVP facial expression detection methods.

\section{Discussions}
Through our systematic literature survey, we have identified a number of factors that will influence the performance of SVP facial expression detection methods. To gain a deeper understanding, we will summarize and discuss these confounding factors as well as some insights and challenges in this section. 

{\it Influence of features.} It is clear that the features extracted for distinguishing between posed and spontaneous facial expressions play a key role in detection performance. Most methods have explored temporal dynamics of different features for effective detection. We can observe from Tables 2-5 that the detection performance varies a lot among different algorithms on the same database. The visual learned features from difference images proposed by \citep{gan2015posed} and \citep{xu2017convolutional} in Table \ref{tab:04} performed better than muscle movement and spatial patterns based methods on SPOS database, while on USTC-NIVE database and smile SVP database UvA-NEMO, spatial patterns based methods achieve slightly higher accuracy than visual features, and significantly higher than other kinds of methods. Overall, visual features based and spatial patterns methods show more promising detection abilities; but there still lacks a consensus about what type of features will be optimal for the task of SVP detection.

{\it Influence of facial regions.}
Each emotion has its own discriminative facial regions, which can be used not only in emotion recognition but also in posed and genuine classification. As mentioned above, study in \citep{zhang2011geometry} has found that in SVP emotion detection, the mouth region is more important for sadness; the nose is more important for surprise; both the nose and mouth regions are important for disgust, fear, happiness, while the eyebrows, eyes, nose, mouth are all important for anger. Another study \citep{liu2012posed} also explored different facial regions, including the forehead, eyes, nose, cheek, and mouth. Experiments results have shown that the forehead and cheek performed better than the other regions for most facial expressions (disgust, fear, sadness, and surprise), while the mouth region performed the worst for most facial expressions. Moreover, fusing all these regions achieved the best performance. In SVP smile detection, it was observed in \citep{dibekliouglu2012you} that the discriminative power of eyelid region is better than cheek and lip corners. A different study in \cite{mandal2017spontaneous} has found that lip-region features (with 73.44\% on UvA-NEMO) outperformed the eye-region features (with 71.14\%), while the combined features performed the best with 74.68\% accuracy. Overall, fusion of multiple facial regions can improve the detection performance over individual features. Besides, varying video temporal segments (i.e., onset, apex, and offset) for feature extraction also lead to different performance. Several studies~\citep{cohn2003timing, dibekliouglu2012you} have demonstrated the onset phase performs best among individual phases in SVP facial expression detection.
 
{\it Influence of emotions.}
Due to the differences in activation of muscles, such as with different intensities and in different facial regions, each emotion has different difficulty levels in SVP expression detection. For example, happiness and anger can activate obvious muscles around eye and mouth regions, which has been widely studied for feature extraction. Based on appearance and geometric features, \citep{zhang2011geometry} found that surprise is the easiest emotion (with 83.4\% accuracy on USTC-NVIE), followed by happiness with 80.5\%,  while disgust is the most difficult one (with 76.1\%). Similarly, \cite{kulkarni2018automatic} achieved better results in detecting SVP happiness (with 71.05\% accuracy) and anger (with 69.40\%), but worse results in disgust (with 63.05\%) and contempt (with 60.85\%) on SASE-FE dataset. On the contrary, \citep{li2017combining} obtained the highest accuracy (of 80\%) for both disgust and happy, while 50\% for contempt on SASE-FE dataset. Overall, SVP happiness is relatively easy to recognize. In the future, how to improve the generalization ability of SVP detection on multiple universal facial expressions, or improve the performance on specific emotion based on its unique facial features, deserves more studies.

{\it Influence of databases.} 
The databases for SVP facial expressions also play a significant role in benchmarking effectiveness and practicality of different detection schemes. From Tables 2-5, it can be observed that the detection performance of the same detection method can vary wildly on different databases. In addition to direct influence of the data size and data quality, studies have also found the influence of subjects in terms of both age and gender. \citep{dibekliouglu2012you} explored the effect of subject age by splitting the UvA-NEMO smile database into young (age $\textless$ 18) and adults (age $\geq$ 18 years), and found that eyelid-and-cheek features provided more reliable classification for adults, while lip-corner features performed better on young people. They further explored the gender effect in method \citep{dibekliouglu2015recognition} and showed that results on males were all better than females using different facial region features. This can be attributed to the reasons that male subjects have more discriminative geometric features (distances between different landmark pairs) than females. They also improved their detection performance by using age or gender as labels. Similarly, \citep{wang2019novel} considered the influence of gender, and incorporated it as privileged information for performance improvement. To sum up, the findings on age and gender influence can not only provide suggestions for SVP facial expression database collection to take subject distribution into consideration, but also inspire researchers to design more effective and practical detection methods taking advantage of data properties.

{\it Influence of classifiers.} 
The classifier has a great effect on most classification tasks, which has also been explored by researchers in  distinction between spontaneous and posed facial expressions. \citep{dibeklioglu2010eyes} assessed the reliability of their features with continuous HMM, k-Nearest Neighbor (k-NN) and naive Bayes classifier, and the highest classification rate was achieved by naive Bayes classifier on two datasets. \citep{pfister2011differentiating} compared support vector machine (SVM), Multiple Kernel Learning (MKL), and Random Forest decision tree (RF) classifier, and found RF outperformed SVM and MKL based on CLBP-TOP features on SPOS database. \citep{dibekliouglu2015recognition} compared Linear Discriminant, Logistic Regression, k-NN, Naive Bayes, and SVM classifiers on UvA-NEMO smile dataset, and showed the outstanding performance of SVM classifier under all testing scenarios. \citep{racovicteanu2019spontaneous} also used SVM, combined with a Hard Negative Mining (HNM) paradigm, to produce the best performance among RF, SVM, and Multi-Layer Perceptron (MLP) classifiers. Overall, as the most widely-used classifier, SVM can provide outstanding performance on several databases. Whether recently developed deep learning-based classifiers can achieve further performance improvement remains to be explored. 

{\it Influence of modalities of images.}
In addition to visible images/videos, some studies have shown the impact of different modalities on improving the detection performance. \citep{pfister2011differentiating}] illustrated that the performance of fusion of NIR with visible images (with 80.0\% accuracy) is better than using single NIR (with 78.2\%) or visible images (with 72.0\%) on SPOS dataset. Although special devices are needed for data acquisition, the advantages of different modalities in revealing subtle features deserve further investigation. It is also plausible to combine the information contained in multiple modalities for detection performance improvement.

{\it Performance differences of spontaneous and posed expressions.}
Last but not the least, several studies have observed the apparent gap of performance  between posed facial expressions detection and genuine ones. For example, based on visual features, \citep{liu2012posed} found that it is much easier to distinguish all posed expressions (with 90.8\% accuracy) than genuine ones (62.6\%) on USTC-NIVE database. Similarly, \citep{mandal2016distinguishing} also achieved higher classification accuracy of posed smiles than spontaneous ones (with over 10\% gaps) on UvA-NEMO dataset. However, two hybrid methods~\citep{mandal2017spontaneous, kulkarni2018automatic} both obtained higher accuracy in detecting genuine facial expressions than posed ones, with a 6\% gap in method \citep{mandal2017spontaneous} on UvA-NEMO Smile database, while an average of 7.9\% gap in method \citep{kulkarni2018automatic} on SASE-FE database. Such  inconsistent differences can be attributed to the influences of both feature extraction methods and databases.

\section{Conclusions}
With the emerging and increasingly supported theory that facial expressions do not always reflect our genuine feelings, automatic detection of spontaneous and posed facial expressions have become increasingly important in human behaviour analysis. This survey has summarized recent advances of SVP facial expression detection over the past two decades. A total of eleven databases and about thirty detection methods have been reviewed and analyzed. Particularly, we have identified and discussed several influencing factors of SVP detection methods to gain a deeper understanding of this nascent field. This review paper is expected to serve as a good starting point for researchers who consider developing automatic and effective models for genuine and posed facial expression recognition.

One area that has not been covered by this survey is the 3D dynamic facial expression databases \citep{zhang2013high,sandbach2012static}. As 3D scanning technology (e.g., Kinect and LIDAR) rapidly advances, SVP detection from 3D instead of 2D data might become feasible in the near future. Can 3D information facilitate the challenging task of SVP detection? It remains to be found out. Research on SVP detection also has connections with other potential applications such as Parkinson's disease \cite{smith1996spontaneous}, deception detection \cite{granhag2004detection}, and alexithymia \cite{mcdonald1990expression}. More sophisticated computational tools such as deep learning based might help boost the research progress in SVP detection. It is likely that the field of facial expression recognition and affective computing will continue growing in the new decade.



%





\begin{acks}
This research was supported by an NSF CAREER Award (1945230), ORAU Ralph E. Powe Junior Faculty Enhancement Award, West Virginia University (WVU), WVU PSCoR Program, and the Dana Foundation (to S.W.), and an NSF Grant (OAC-1839909) and the WV Higher Education Policy Commission Grant (HEPC.dsr.18.5) (to X.L.). The funders had no role in study design, data collection and analysis, decision to publish, or preparation of the manuscript.
\end{acks}

\end{document}